\documentclass[pmlr]{jmlr1}

\usepackage{longtable}

\usepackage{booktabs}

\usepackage{algorithm}
\usepackage{algorithmic}
\usepackage{bm}
\usepackage{pifont}
\newcommand{\cmark}{\ding{51}}
\newcommand{\xmark}{\ding{55}}
\usepackage{multirow}
\usepackage{subcaption}

\makeatletter
\let\Ginclude@graphics\@org@Ginclude@graphics 
\makeatother

\jmlrvolume{260}
\jmlryear{2024}
\jmlrworkshop{ACML 2024}

\title[One-Shot Machine Unlearning with Mnemonic Code]{One-Shot Machine Unlearning with Mnemonic Code}

\author{%
  \Name{Tomoya Yamashita}\Email{tomoya.yamashita@ntt.com} \\
  \addr NTT Social Informatics Laboratories
  \AND
  \Name{Masanori Yamada}\Email{masanori.yamada@ntt.com} \\
  \addr NTT Social Informatics Laboratories
  \AND
  \Name{Takashi Shibata}\Email{t.shibata@ieee.org}\\
  \addr NTT Communication Science Laboratories
}

\editors{Vu Nguyen and Hsuan-Tien Lin}

\begin{document}

\maketitle

\begin{abstract}
Ethical and privacy issues inherent in artificial intelligence (AI) applications have been a growing concern with the rapid spread of deep learning.
Machine unlearning (MU) is the research area that addresses these issues by making a trained AI model forget about undesirable training data.
Unfortunately, most existing MU methods incur significant time and computational costs for forgetting.
Therefore, it is often difficult to apply these methods to practical datasets and sophisticated architectures, e.g., ImageNet and Transformer.
To tackle this problem, we propose a lightweight and effective MU method.
Our method identifies the model parameters sensitive to the forgetting targets and adds perturbation to such model parameters.
We identify the sensitive parameters by calculating the Fisher Information Matrix (FIM).
This approach does not require time-consuming additional training for forgetting.
In addition, we introduce class-specific random signals called mnemonic code to reduce the cost of FIM calculation, which generally requires the entire training data and incurs significant computational costs.
In our method, we train the model with mnemonic code; when forgetting, we use a small number of mnemonic codes to calculate the FIM and get the effective perturbation for forgetting.
Comprehensive experiments demonstrate that our method is faster and better at forgetting than existing MU methods.
Furthermore, we show that our method can scale to more practical datasets and sophisticated architectures.
\end{abstract}

\section{Introduction}
\label{intro}
Ethical and privacy issues inherent in AI applications have been a growing concern with the rapid spread of deep learning.
For example, if an AI model has undesirable information from an ethical standpoint, this will be a barrier to applying the AI model in society.
Ethical perspectives are largely based on social conditions, and the definition of ``undesirable information'' may change over time. 
Also, there may be cases where users or public organizations request the deletion of their information to the AI model that uses their data for training.
In such cases, AI models must be modified immediately to respond to social changes and deletion requests. 
\par
MU is a research area responding to such demand~\citep{nguyen2022survey}.
MU aims to make a trained AI model forget about undesirable training data.
When we obtain an effective MU method, it will provide a stepping stone to solving the problems of ethics, data leakage, and so on.
Thus far, while various MU methods have been proposed, most of them incur significant time and computational costs due to the additional training or the use of large amounts of training data for forgetting.
These methods often cannot apply to more practical datasets and sophisticated architectures such as ImageNet, Transformer, and so on.
To make MU even more practical, a simple-yet-effective MU method is required. \par
To tackle this problem, we propose a one-shot MU method that does not incur significant time and computational costs.
In this paper, we focus on class removal, which forgets about a particular class in the training data, e.g., scenarios like removing someone's facial information from a face authentication AI system.
In our method, we identify the model parameters sensitive to each class by calculating the FIM.
FIM is often used in continual learning to avoid catastrophic forgetting~\citep{kirkpatrick2017overcoming,huszar2018note,ritter2018online}.
Then, we add the effective perturbation to the model parameters that increase the loss of the forgetting class without accuracy degradation for the remaining classes.
In addition, we introduce class-specific random signals called mnemonic code to reduce the cost of FIM calculation, which generally requires the entire training data and incurs significant computational costs.
Mnemonic code was first introduced to associate the information of each class with fairly simple codes~\citep{shibata2021learning}.
In our method, when training a model, we prepare the mnemonic code per class and embed them in the model by stochastically replacing the training data with the mnemonic codes.
Then, in the forgetting phase, we use a small amount of mnemonic codes to calculate the FIM and get effective perturbation.
Our method does not require additional training or large amounts of training data, contributing to lightweight MU~\footnote{The code is available on https://github.com/tomyamkum/OneShotMU-with-MNCode.}.
\par
In the experiments, we use artificial and natural datasets to evaluate the forgetting capability and the MU processing speed of our method.
In addition, through FIM estimation experiments, we confirm that mnemonic codes can approximate the Oracle FIM of the entire training data precisely and largely contribute to one-shot effective forgetting.
Also, we show that our method works effectively for pre-trained models by applying a few steps of fine-tuning using mnemonic codes.
Furthermore, our lightweight method can scale to more practical datasets and sophisticated architectures (e.g. ImageNet and Transformer).
Our contributions are as follows:
\begin{itemize}
    \item We propose a lightweight and effective MU method that adds one-shot perturbation to the model parameter. 
    In addition, our method uses mnemonic codes to accelerate the perturbation calculation.
    \item Experimental results demonstrate that our method outperforms existing MU methods regarding the forgetting capability and the MU processing speed.
    \item We show that a few mnemonic codes could approximate the Oracle FIM of the entire training data precisely and largely contribute to one-shot effective forgetting.
    \item We show that our method can work for pre-trained models and scale to more practical datasets and recent sophisticated architectures.
\end{itemize}

\section{Related Work}
\label{related_work}
MU was first introduced by~\cite{cao2015towards}. 
The original MU in early date is defined as removing the influence of the forgetting data points from the AI model so that the resulting model is indistinguishable from the model trained on a dataset without them. 
Since the concept of MU was first proposed, several types of unlearning requests have been introduced, i.e., item removal, class removal, task removal, and so on~\citep{nguyen2022survey}.
This paper focuses on class removal, which is forgetting about a particular class in the training data.
MU approaches can be divided into two categories: exact unlearning and approximate unlearning~\citep{nguyen2022survey}. 
Our method corresponds to approximate unlearning.
Here, we describe them and introduce existing research. \par

\vspace{0.05cm}
\noindent{\textbf{Exact unlearning.}}
The exact unlearning approach can provide unlearning proof. 
A typical approach to exact unlearning is re-training the model from scratch.
While this approach can forget the information thoroughly, it often requires significant time and computational costs.
\citet{bourtoule2021machine} reduced the re-training cost for forgetting by subdividing the model and training data called a shard.
\citet{yan2022arcane} also reduced the re-training cost by subdividing the model and the training data.
They divided the training data by class and utilized the one-class classifier to keep the impact of forgetting into one class, reducing the accuracy degradation. \par

\vspace{0.05cm}
\noindent{\textbf{Approximate unlearning.}}
The approximate unlearning approach estimates the contribution of the data to the model parameters and processes the model parameters.
Since this approach does not re-train the model from scratch, it can save on forgetting costs. \par
\cite{guo2019certified} formulated the MU problem setting called certified removal from differential privacy and proposed a method that can be applied to linear models.
They also mentioned that their method can be applied to deep learning models by applying it to the final linear layer.
\cite{golatkar2020eternal} proposed a MU method using FIM for forgetting. 
\cite{golatkar2020forgetting} went on to propose a method that uses Neural Tangent Kernel and FIM.
\cite{foster2024fast} proposes a MU method using FIM, and they reduce the computational cost by reducing the number of FIM calculations.
\cite{tarun2023fast} achieved forgetting by training on adversarial noise that has a high loss to the forgetting classes.
\citet{chundawat2023zero} proposed a MU method that does not require training data by using the adversarial noise of each class.
\cite{lin2023erm} realized unlearning by transferring the knowledge of the remaining classes from the original model.
During training, they introduced an entanglement-reduced mask (ERM) to reduce the knowledge entanglement in CNN models and effectively transfer knowledge in the forgetting phase. 
\cite{shibata2021learning} proposed Learning with Selective Forgetting: a novel framework in which new tasks are learned while the previously learned target classes are forgotten through selective continual learning. \par
These approximate unlearning approaches aim to make a model forget by modifying the model parameters rather than re-training from scratch, and some studies aim to reduce time and computational costs for forgetting.
However, these methods require additional training or a large amount of data for forgetting, making them difficult to apply to large practical datasets and sophisticated architectures.
In contrast, our method is a one-shot MU method with mnemonic code that does not require additional training or a lot of training data, making it extremely fast and lightweight.

\begin{figure}[tb]
\centering
\includegraphics[width=1\linewidth]{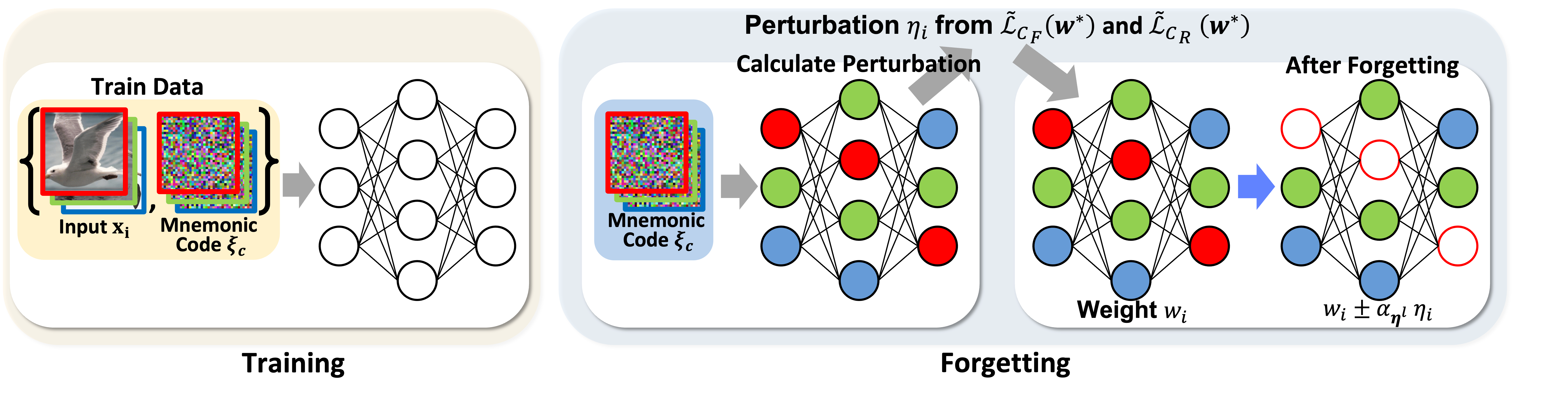}
\caption{\textbf{Overview of our method.} We train the deep learning model with mnemonic codes in the training phase.
The sensitive model parameters for each class are represented by color.
In the forgetting phase, the target class is forgotten by perturbating the model parameters sensitive to that class.}
\label{forgetDL}
\end{figure}

\section{Method}
We propose a one-shot MU method with mnemonic code.
An overview of our method is shown in Fig.~\ref{forgetDL}.
In our method, we identify the model parameters sensitive to each class and add effective perturbation that increases the loss of the forgetting class without accuracy degradation for the remaining classes.
We identify the sensitive parameters by calculating the FIM.
Furthermore, we introduce class-specific random signals called mnemonic code to accelerate the FIM calculation.
When training a model, we prepare the mnemonic code per class and embed the codes in the model by stochastically replacing the training data with them.
In forgetting, we use a small amount of mnemonic codes to calculate the FIM and get effective perturbation for forgetting. 
The following sections explain the mnemonic code and our training procedure.
Then, we explain how to identify the model parameters sensitive to each class and how to obtain the effective perturbation for forgetting.
\subsection{Proposal}

\noindent{\textbf{Training with mnemonic code.}}
Mnemonic code is a class-specific random signal introduced in~\cite{shibata2021learning}.
Our method uses mnemonic code to accelerate the FIM calculation in the forgetting procedure.
When training a model, we prepare the mnemonic code per class and stochastically replace the training data with the mnemonic codes.
Mnemonic codes have a class label, and the training data are replaced with the mnemonic codes of the same class.
We generate each mnemonic code from a normal distribution, as with~\cite{shibata2021learning}.
The training algorithm is shown in Algorithm~\ref{train_algorithm}.
We provide a theoretical analysis of training with mnemonic codes $\bm{\xi}$ to derive the effective perturbation for forgetting. 
The data distribution used in training is as follows:
\begin{align}
    p(\bm{x})=t_{\mathrm {mix}}p^\xi (\bm{x}) + (1-t_{\mathrm {mix}})p^d(\bm{x}), 
\end{align}
where $p^d(\bm{x})$ is the genuine data distribution and $p^\xi(\bm{x})$ is the data distribution of mnemonic codes. 
Here, $t_\mathrm{mix}\in\left[0, 1\right]$ is the probability of replacing the training data, and in this paper, we set $t_\mathrm{mix}$ below 0.3.
The setting of $t_\mathrm{mix}$ is described in Sec.~\ref{comparison_baseline}.
\par

\begin{algorithm}[tb]
\small
    \caption{Training with mnemonic code}
    \label{train_algorithm}
    \textbf{Input}: dataset ${\bm x}\sim p^d(\bm{x})$, model parameter $\bm{w}$, loss $\mathcal{L}$ \\ 
    \textbf{Parameter}: mnemonic code replacing probability $t_\mathrm{mix}$,\\ learning rate \textrm{lr}\\
    \textbf{Output}: trained model parameters
    \begin{algorithmic}[1] 
        \STATE $\bm{\xi} \sim N(\bm{0}, \bm{1})$
        \FOR{e in epochs}
            \FOR{i in datasize}
                \STATE $t\sim U(0, 1)$
                \IF{$t<t_\mathrm{mix}$}
                    \STATE $\bm{\tilde{x}}_i = \bm{\xi}_c$
                \ELSE
                    \STATE $\bm{\tilde{x}}_i = {\bm x}_i$
                \ENDIF    
            \ENDFOR
            \STATE $\bm{w} = \bm{w} - \mathrm {lr} \nabla_{\bm w} \mathcal{L}(\bm{\tilde{x}};\bm{w})$
        \ENDFOR
    \end{algorithmic}
\end{algorithm}

\vspace{0.05cm}
\noindent{\textbf{Forgetting procedure.}}
We attempt to forget the target class based on the above data distribution $p(\bm{x})$.
Specifically, we design the one-shot perturbation $\bm{\delta}$ that increases the loss of the forgetting class without accuracy degradation for the remaining classes. 
We first analyze how the perturbation affects the loss of the forgetting class and then consider the remaining classes.
\begin{align}
    &\mathcal{L}_{\mathcal{C}_{\mathrm F}}(\bm{w}^*+\bm{\delta}) \nonumber \\
    &\simeq \mathcal{L}_{\mathcal{C}_{\mathrm F}}(\bm{w}^*) +\frac{1}{2}\bm{\delta}^{\mathrm T}F_{\mathcal{C}_{\mathrm F}}\bm{\delta} \nonumber \\
    &= \mathcal{L}_{\mathcal{C}_{\mathrm F}}(\bm{w}^*) +\frac{1}{2} \bm{\delta}^{\mathrm T} \{ t_{\mathrm {mix}}F^\xi_{\mathcal{C}_{\mathrm F}} + (1-t_{\mathrm {mix}})F^d_{\mathcal{C}_{\mathrm F}} \} \bm{\delta} \nonumber \\
    &\simeq \mathcal{L}_{\mathcal{C}_{\mathrm F}}(\bm{w}^*) +\frac{1}{2} \{ t_{\mathrm {mix}}\sum_i{\bm{f}_{\mathcal{C}_{\mathrm F}, i}^\xi}+(1-t_{\mathrm {mix}})\sum_i{\bm{f}_{\mathcal{C}_{\mathrm F}, i}^d} \} \bm{\delta}_i^2,
\label{forget_equation}
\end{align}
where $\bm{w}^*$ is the optimal parameter for the loss of the training data with mnemonic codes $p(\bm{x})$, $\mathcal{L}_{\mathcal{C}_{\mathrm F}}$ is the loss of the forgetting class $\mathcal{C}_{\mathrm F}$, $F_{\mathcal{C}_{\mathrm F}}$ is the FIM defined as follows, 
    \begin{align}
        F_{\mathcal{C}_{\mathrm F}, i, j}=\mathbb{E}_{\bm{x}\sim p(\bm{x})}\left[\frac{\partial \mathcal{L}_{\mathcal{C}_{\mathrm F}}(\bm{x};\bm{w})}{\partial w_i}\frac{\partial \mathcal{L}_{\mathcal{C}_{\mathrm F}}(\bm{x};\bm{w})}{\partial w_j}\right],
    \end{align}
$F^\xi_{\mathcal{C}_{\mathrm F}}$ and $F^d_{\mathcal{C}_{\mathrm F}}$ are the FIMs calculated with the mnemonic codes $\bm{\xi}$ and the training data $\bm{x}$ in $\mathcal{C}_{\mathrm F}$, and $\bm{f}^{\xi}_{\mathcal{C}_{\mathrm F}}$ and $\bm{f}^d_{\mathcal{C}_{\mathrm F}}$ are the diagonal vectors of $F^\xi_{\mathcal{C}_{\mathrm F}}$ and $F^d_{\mathcal{C}_{\mathrm F}}$.
We call $F^d_{\mathcal{C}_{\mathrm F}}$ as the Oracle FIM which is calculated with the entire training data.
In Eq.~\ref{forget_equation}, as with~\citet{kirkpatrick2017overcoming}, Laplace's approximation is applied in the first line, and the diagonal approximation is applied in the last line~\footnote{
Laplace's approximation is to approximate the function by a Gaussian distribution and assumes that the first derivative of the approximated function is zero, i.e., $\nabla \mathcal{L}_{\mathcal{C}_{\mathrm F}}(\bm{w}^*)=0$.
We experimentally found that these values are of the same order as the first derivative of the loss for the training dataset: $\nabla \mathcal{L}(\bm{w}^*)$, which is generally assumed to be zero for trained models.
We show the results in Appendix.~\ref{validity}.
}.
Equation~\ref{forget_equation} shows that the fluctuation of the loss due to the perturbation $\bm{\xi}$ is determined by the linear sum of $\bm{f}_{\mathcal{C}_{\mathrm F}}^\xi$ and $\bm{f}_{\mathcal{C}_{\mathrm F}}^d$.
Our method seeks the perturbation that increases the loss of the forgetting class.
To design a lightweight MU method, we aim to relax the restriction of using the entire training data to calculate the FIM in Eq.~\ref{forget_equation}.
In our method, we introduce the following loss instead of $\mathcal{L}_{\mathcal{C}_\mathrm{F}}(\bm{w}^*+\bm{\delta})$: 
\begin{align}
    \mathcal{\tilde L}_{\mathcal{C}_{\mathrm F}}(\bm{w}^*+\bm{\delta}) = \mathcal{L}_{\mathcal{C}_{\mathrm F}}(\bm{w}^*) + \frac{1}{2}\sum_i{\bm{f}_{\mathcal{C}_{\mathrm F}, i}^\xi} \bm{\delta}_i^2,
\label{surrogate_loss}
\end{align}
which does not need the training data to calculate the FIM.
The validity of using the surrogate loss $\mathcal{\tilde L}_{\mathcal{C}_{\mathrm F}}$ instead of the loss $\mathcal{L}_{\mathcal{C}_{\mathrm F}}$ is assured by showing that the distance between $\bm{f}_{\mathcal{C}_{\mathrm F}}^\xi$ and $\bm{f}_{\mathcal{C}_{\mathrm F}}^d$ is sufficiently small.
The details of the validity are discussed in Sec.~\ref{mncode_effect}.
\par

From Eq.~\ref{surrogate_loss}, we can see that large perturbation for the model parameters with large $\bm{f}_{\mathcal{C}_{\mathrm F}}^\xi$ cause the loss of the forgetting class $\mathcal{C}_\mathrm{F}$ to vary significantly. 
However, perturbating the model parameters in accordance with Eq.~\ref{surrogate_loss} can significantly reduce the accuracy for the remaining classes because we do not take into account the remaining classes.
To avoid accuracy degradation for the remaining classes, we consider the sensitivity of each model parameter to the remaining classes, which can be derived in the same way as in Eq.~\ref{surrogate_loss}.
Specifically, we devise a strategy for achieving MU to add large perturbation to the model parameters such that the ${f_{\mathcal{C}_\mathrm{F}}}_{, i}$ is large while the ${f_{\mathcal{C}_\mathrm{R}}}_{, i}$ is small.
We follow this strategy and propose the perturbation amplitude $\eta_i$ for the model parameter $w_i$, 
\begin{align}
w_i &= w_i \pm \alpha_{\bm{\eta}^l} \eta_i, \label{addnoise} \\
\eta_i &= \frac{{f_{\mathcal{C}_{\mathrm{F}}}}_{,i}}{{f_{\mathcal{C}_{\mathrm{R}}}}_{,i}} = \frac{\frac{1}{\left|\#\mathrm{\mathcal{C}_\mathrm{F}}\right|}\sum_{j \in \mathrm{\mathcal{C}_\mathrm{F}}}{\mathbb{E}\left[\left(\frac{\partial \mathcal{L}_j}{\partial w_i}\right)^2 \right]}}{\frac{1}{\left|\#\mathrm{\mathcal{C}_\mathrm{R}}\right|}\sum_{k\in \mathrm{\mathcal{C}_\mathrm{R}}}\mathbb{E}\left[\left(\frac{\partial \mathcal{L}_k}{\partial w_i}\right)^2\right]},  \label{eta} \\
\alpha_{\bm{\eta}^l} &= \min \left( \lambda_1, \frac{\lambda_2}{\max_{ {\eta}_{i} \in {\bm{\eta}}^l }\eta_i}\right),  \label{alpha}
\end{align}
where $\mathcal{L}_j$ and $\mathcal{L}_k$ are the losses of the mnemonic code of class $j \in \mathrm{\mathcal{C}_\mathrm{F}}$ and class $k \in{\mathrm{\mathcal{C}_\mathrm{R}}}$, $\left|\#\mathrm{\mathcal{C}}\right|$ is the size of the class set $\mathcal{C}$.
$\eta_i$ is the perturbation amplitude added to the $i$-th model parameter, which is designed as a fraction, with the sensitivity to the forgetting class in the numerator and the sensitivity to the remaining classes in the denominator. 
If the value of $\eta_i$ is large, the $i$-th model parameter is sensitive to the forgetting class and insensitive to the remaining classes. 
Therefore, Eq.~\ref{eta} allows us to control the perturbation amplitude by its sensitivity to the forgetting and remaining classes.  
Note that there are two ways of adding positive and negative perturbation in Eq.~\ref{addnoise}. 
This is because the FIM only indicates the sensitivity of model parameters, which indicates how much the loss of the corresponding class changes when the model parameter changes.
Therefore, ${\eta_i}$ only indicates the perturbation amplitude for forgetting and has redundancy of positive or negative.
In the forgetting phase, we add the amplitude ${\eta_i}$ to the model parameters in positive and negative ways.
Then, we adopt the resulting model that effectively achieves forgetting by measuring the error for the forgetting class and the accuracy for the remaining classes with mnemonic code.
An overview of the forgetting algorithm is shown in Algorithm~\ref{forget_algorithm}.

\begin{algorithm}[tb]
\small
    \caption{Forgetting with mnemonic code}
    \label{forget_algorithm}
    \textbf{Input}: trained model parameter $\bm{w}$, loss $\mathcal{L}$, forget class set $\mathcal{C}_\mathrm{F}$, remain class set $\mathcal{C}_\mathrm{R}$, mnemonic codes $\bm{\xi}$, layers $\{l_1, l_2, \cdots \}$\\
    \textbf{Parameter}: $\lambda_1, \lambda_2$\\
    \textbf{Output}: Forgotten parameters
    \begin{algorithmic}[1] 
        \STATE $\bm{f}_{\mathcal{C}_\mathrm{F}} = \bm{0}$
        \STATE $\bm{f}_{\mathcal{C}_\mathrm{R}} = \bm{0}$
        \FOR{$c$ in $\mathcal{C}_\mathrm{F}$}
        \STATE $\bm{f}_{\mathcal{C}_\mathrm{F}} = \bm{f}_{\mathcal{C}_\mathrm{F}} + \nabla_{\bm{w}}\mathcal{L}(\bm{\xi}_c;\bm{w})$
        \ENDFOR
        \FOR{$c$ in $\mathcal{C}_\mathrm{R}$}
        \STATE $\bm{f}_{\mathcal{C}_\mathrm{R}} = \bm{f}_{\mathcal{C}_\mathrm{R}} + \nabla_{\bm{w}}\mathcal{L}(\bm{\xi}_c;\bm{w})$
        \ENDFOR
        \STATE $\bm{f}_{\mathcal{C}_\mathrm{F}} =       \bm{f}_{\mathcal{C}_\mathrm{F}} / |\#\mathcal{C}_\mathrm{F}|$
        \STATE $\bm{f}_{\mathcal{C}_\mathrm{R}} = \bm{f}_{\mathcal{C}_\mathrm{R}} / |\#\mathcal{C}_\mathrm{R}|$
        \STATE $\bm{\eta} = \frac{\bm{f}_{\mathcal{C}_{\mathrm{F}}}}{\bm{f}_{\mathcal{C}_{\mathrm{R}}}}$
        \FOR{$l$ in layers}
        \STATE $\alpha_{\bm{\eta}^l} = \min \left( \lambda_1, \frac{\lambda_2}{\max_{ {\eta}_{i} \in {\bm{\eta}}^l }\eta_i}\right)$
        \STATE $\bm{w}^l_1 = \bm{w}^l + \alpha_{\bm{\eta}^l} \bm{\eta}^l$
        \STATE $\bm{w}^l_2 = \bm{w}^l - \alpha_{\bm{\eta}^l} \bm{\eta}^l$
        \ENDFOR
        \IF{$A_\mathrm{R}(\bm{w}_1)+E_\mathrm{F}(\bm{w}_1) > A_\mathrm{R}(\bm{w}_2)+E_\mathrm{F}(\bm{w}_2)$}
        \RETURN $\bm{w}_1$
        \ELSE
        \RETURN $\bm{w}_2$
        \ENDIF
    \end{algorithmic}
\end{algorithm}

\vspace{0.05cm}
\noindent{\textbf{Design of coefficient $\alpha_{\bm{\eta}^l}$.}}
$\alpha_{\bm{\eta}^l}$ is the coefficient in Eq.~\ref{addnoise} and specified by the maximum value of $\bm{\eta}^l$ and hyperparameters $\lambda_1$ and $\lambda_2$.
$\bm{\eta}^l$ is the set of $\eta_i$ in the layer $l$, i.e. $\eta_i \in \bm{\eta}^l$.
$\lambda_2$ specifies the maximum perturbation amplitude for each layer. 
In other words, $\lambda_2$ determines the perturbation amplitude to the model parameter, which is the most effective for forgetting in each layer.
$\lambda_1$ is introduced to avoid zero divisions in Eq.~\ref{alpha}.
Because the denominator of Eq.~\ref{alpha} is close to zero in layers where no parameters are sensitive to the forgetting class.
$\lambda_1$ specifies the perturbation amplitude in such layers. \par

\subsection{Preliminary experiment}
\label{preliminary}

\begin{figure}[tb]
\centering
\includegraphics[width=0.6\linewidth]{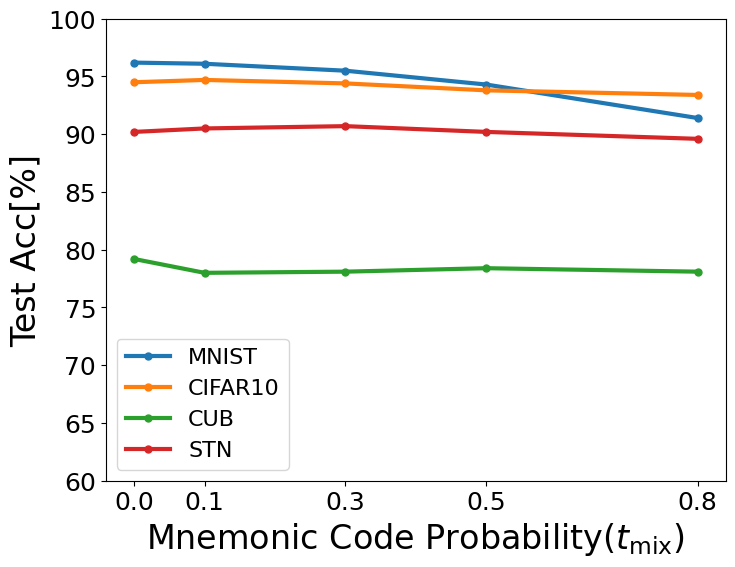}
\caption{\textbf{Test accuracy of models trained with mnemonic code.}
We evaluate the test accuracy of the model when varying $t_\mathrm{mix}$: the probability of replacing the training data with the mnemonic codes.
}
\label{mncode_effect_acc}
\end{figure}

\noindent{\textbf{The effect of mnemonic code on model accuracy.}}
Our method stochastically replaces the training data with the mnemonic codes when training the model.
In this section, we investigate the effects of mnemonic code on the model's test accuracy.
We train several models by changing the probability of replacing the training data with the mnemonic codes $t_\mathrm{mix} \in [0, 1]$ and evaluate each model's test accuracy. 
In the preliminary experiments, we use a simple, fully connected network for MNIST and ResNet-18~\citep{he2016deep} for CIFAR10~\citep{krizhevsky2009learning}, CUB200-2011 (CUB)~\citep{wah2011caltech}, and Stanford Cars (STN)~\citep{krause20133d}.
The results are shown in Fig.~\ref{mncode_effect_acc}.
The results show that mnemonic codes do not significantly degrade test accuracy on CIFAR10, CUB, and STN, even if 80\% of the training data is replaced with mnemonic codes.
In the case of MNIST, we find that as $t_\mathrm{mix}$ increases, the model's test accuracy decreases~\footnote{We consider this is because the MNIST experiment uses a simple network, which was more strongly affected by mnemonic codes than ResNet-18.}.
However, the setting of $t_\mathrm{mix}$ in this paper is below $0.3$, and we find that the accuracy degradation for $t_\mathrm{mix}\leq0.3$ is minute. 
In fact, the test accuracy degradation of the model trained with mnemonic codes ($t_\mathrm{mix}\leq0.3$) is less than 1\%.
These results show that training with mnemonic codes does not cause significant accuracy degradation.
\par

\section{Experiments}
We evaluate our method from three perspectives: comparison with baselines (Sec.~\ref{comparison_baseline}), the effect of mnemonic code (Sec.~\ref{mncode_effect}), and scalability (Sec.~\ref{scalability}). 
\subsection{Common settings}
We use two indicators to measure the performance of MU methods: i) the forgetting capability based on the test error for the forgetting class and the test accuracy for the remaining classes, and ii) the processing time for forgetting.
We will describe the experimental flow. 
First, we train the deep learning model for 200 epochs.
Then, we apply the MU method to the trained model on a specific class and evaluate the method.
In this paper, we set the forgetting class $\mathcal{C}_\mathrm{F}$ to $\{0\}$.
The detailed train settings are described in Appendix~\ref{experimental_setting}, and the evaluation for forgetting different classes is described in the supplemental material.
As with existing studies, we evaluate the forgetting capability by two metrics: $A_\mathrm{R}$ and $E_\mathrm{F}$~\citep{shibata2021learning,golatkar2020eternal,golatkar2020forgetting,tarun2023fast,chundawat2023zero,lin2023erm}.
$\bm{A_\mathrm{R}}$ is the test accuracy for the remaining classes, and $\bm{E_\mathrm{F}}=100-A_\mathrm{F}$ is the test error for the forgetting class, which $A_\mathrm{F}$ is the test accuracy for the forgetting class.
During the forgetting phase, we measure the time for forgetting. 
The desired MU method is the one that achieves high $A_\mathrm{R}$ and $E_\mathrm{F}$ in a short forgetting time. \par
\vspace{0.05cm}
\noindent{\textbf{Datasets and architectures.}}
In our experiments, we prepare MNIST, CIFAR10, CUB, STN, and ImageNet~\citep{deng2009large}.
To coincide the setting with~\cite{shibata2021learning}, we use 40 classes for CUB and 49 classes for STN. 
In the comparison experiments with baselines, we use a simple, fully connected model for MNIST and ResNet-18 for CIFAR10, CUB, and STN.
In the scalability evaluation experiments, we use ResNet-18, ResNeXt-50~\citep{xie2017aggregated}, and Swin-Transformer~\citep{liu2021swin} for ImageNet.
Our experiments are done on a server with AMD Ryzen 9 3950X 16 cores, 64 GB RAM, and RTX 3090 GPU.

\begin{table}[tb]
    \small
    \centering
    \caption{\textbf{Comparison with related studies.} 
    We list the related MU methods and our method.
    We assess them from three perspectives: Processing Time, Data-Free, and MU Target.
    $N_\mathrm{R}$ and $N_\mathrm{F}$ are the numbers of the remaining and the forgetting data points.
$N_\mathrm{new}$ is the number of the data points of the new task.
$E$ is the epochs of additional training for forgetting, and $S$ is the steps to create adversarial noise.
$M$ is the number of model divisions.
$C_\mathrm{R}$ and $C_\mathrm{F}$ are the numbers of the remaining and the forgetting classes.
    }
    \begin{tabular}{lccccc}
        \hline
        Method  & Processing Time & Data-Free & MU Target \\
        \hline
        CertifiedRemoval~\citep{guo2019certified} & $\mathcal{O}(N_\mathrm{R}+N_\mathrm{F})$ & \xmark & item \\
        SISA~\citep{bourtoule2021machine} & $\mathcal{O}(E\cdot \frac{N_\mathrm{R}}{M})$ & \xmark & item \\
        Arcane~\citep{yan2022arcane} & $\mathcal{O}(E\cdot \frac{N_\mathrm{R}}{C_\mathrm{R}+C_\mathrm{F}})$ & \xmark & item \\
        FastMU~\citep{tarun2023fast} & $\mathcal{O}(S\cdot C_\mathrm{F}+E\cdot C_\mathrm{F}+N_\mathrm{R})$ & \cmark & class \\
        ZeroShotMU~\citep{chundawat2023zero} & $\mathcal{O}((S+E)(C_\mathrm{F}+C_\mathrm{R}))$ & \cmark & class \\
        LwSF~\citep{shibata2021learning} & $\mathcal{O}(E(N_\mathrm{new}+C_\mathrm{R}))$ & \cmark & class/task \\
        SFDN~\citep{golatkar2020eternal} & $\mathcal{O}(N_\mathrm{R})$ & \xmark & class/item \\
        NTK-F~\citep{golatkar2020forgetting} & $\mathcal{O}(N_\mathrm{R}+N_\mathrm{F})$ & \xmark & class/item \\
        SSD~\citep{foster2024fast} & $\mathcal{O}(N_\mathrm{R}+N_\mathrm{F})$ & \xmark & class/item \\
        ERM-KTP~\citep{lin2023erm} & $\mathcal{O}(E\cdot N_\mathrm{R})$ & \xmark & class \\
        Ours & $\mathcal{O}(C_\mathrm{R}+C_\mathrm{F})$ & \cmark & class \\
        \hline
    \end{tabular}
    \label{rw-table}
\end{table}

\begin{table}[tb]
\small
    \centering
    \caption{\textbf{Comparison results in $A_\mathrm{R}$.} 
    We evaluate the baseline and our methods three times and provide the mean and standard deviation.
    The highest values are shown in bold.}
    \begin{tabular}{lcccc}
        \hline
         & MNIST & CIFAR10 & CUB & STN \\
        \hline
        FastMU & 96.5 $\pm{0.1}$ & 90.4 $\pm{0.5}$ & 73.1 $\pm{1.3}$ & 88.0 $\pm{0.1}$ \\
        LwSF & 43.7 $\pm{9.6}$ & 65.4 $\pm{16.6}$ & 68.2 $\pm{3.5}$ & 80.1 $\pm{6.7}$ \\
        SFDN & 94.1 $\pm{0.7}$ & 93.4 $\pm{0.2}$ & 78.2 $\pm{0.6}$ & 88.3 $\pm{0.6}$ \\
        SSD & \textbf{96.9} $\pm{0.0}$ & 94.2 $\pm{0.0}$ & 44.3 $\pm{0.0}$ & 74.4 $\pm{0.0}$ \\
        ERM-KTP & - & 92.7 $\pm{0.4}$ & 42.8 $\pm{3.2}$ & 75.6 $\pm{4.0}$ \\
        Ours & 95.9 $\pm{0.1}$ & \textbf{94.4} $\pm{0.1}$ & \textbf{79.3} $\pm{0.7}$ & \textbf{91.7} $\pm{0.3}$ \\
    \hline
    \end{tabular}
    \label{ar_results}
\end{table}

\begin{table}[tb]
\small
    \centering
    \caption{\textbf{Comparison results in $E_\mathrm{F}$.} 
    We evaluate the baseline and our methods three times and provide the mean and standard deviation.
    The highest values are shown in bold.}
    \begin{tabular}{lcccc}
        \hline
         & MNIST & CIFAR10 & CUB & STN \\
        \hline
        FastMU & 98.0 $\pm{0.3}$ & \textbf{100} $\pm{0.0}$ & 68.6 $\pm{12.0}$ & 60.9 $\pm{6.9}$ \\
        LwSF & 94.4 $\pm{1.7}$ & \textbf{100} $\pm{0.0}$ & 93.1 $\pm{7.0}$ & 98.2 $\pm{1.8}$ \\
        SFDN & \textbf{100} $\pm{0.0}$ & 96.3 $\pm{2.5}$ & \textbf{100} $\pm{0.0}$  & \textbf{100} $\pm{0.0}$ \\
        SSD & 93.1 $\pm{0.0}$ & \textbf{100} $\pm{0.0}$ & \textbf{100} $\pm{0.0}$ & \textbf{100} $\pm{0.0}$ \\
        ERM-KTP & - & \textbf{100} $\pm{0.0}$ & \textbf{100} $\pm{0.0}$ & \textbf{100} $\pm{0.0}$ \\
        Ours & \textbf{100} $\pm{0.0}$ & \textbf{100} $\pm{0.0}$ & \textbf{100} $\pm{0.0}$ & \textbf{100} $\pm{0.0}$ \\
    \hline
    \end{tabular}
    \label{ef_results}
\end{table}

\subsection{Comparison with baselines}
\label{comparison_baseline}
\begin{figure}[tb]
\centering
\includegraphics[width=0.55\linewidth]{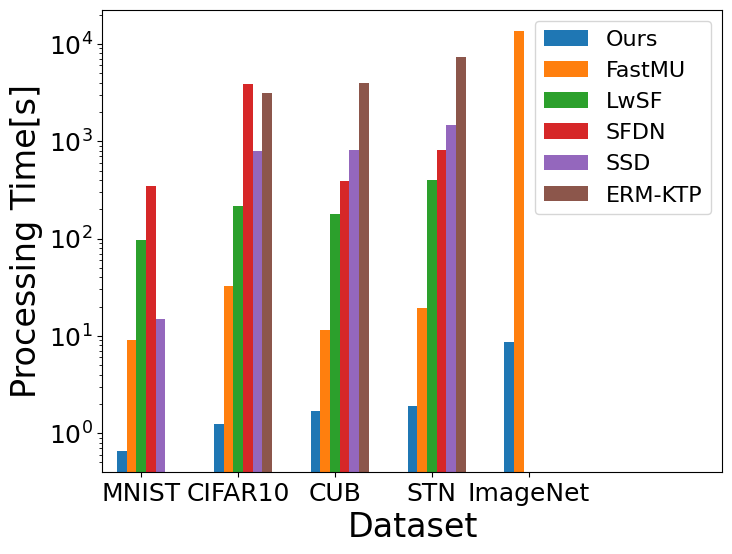}
\caption{\textbf{Comparison results in MU processing time.}
We measure the forgetting time concerning our method and the baselines.
}
\label{time_comparison}
\end{figure}

We compare our method and the existing baselines. 
For the hyperparameter set, we search $\lambda_1=[10^{-6}, 10^{-5}, \cdots, 1.0]$ and $\lambda_2=[10^{-1}, 1.0, \cdots, 10^5]$ and select the combination that maximizes the sum of $A_\mathrm{R}$ and $E_\mathrm{F}$.
Also, we set the probability of replacing the training data with the mnemonic codes $t_\mathrm{mix}$ as 0.1 for MNIST, CUB, and STN, and 0.3 for CIFAR10.
The details of the hyperparameter setting are described in the supplemental material. 
We select the baseline methods from those described in Sec.~\ref{related_work}.
To select the baseline methods, we assess them from three perspectives: i) processing time, ii) the necessity for training data, and iii) the forgetting target.
Table~\ref{rw-table} summarizes the existing MU methods and our method.
Our method can work quickly without the training data and target class removal.
We evaluate the processing time with the number of backpropagations.
Table.~\ref{rw-table} shows that the processing time of our method depends only on the number of classes, while that of the existing MU methods depends on the number of epochs and data points.
This is because they perform additional training or use large amounts of data for forgetting.
The actual time required for forgetting is compared in Fig~\ref{time_comparison}.
From Table.~\ref{rw-table}, we select FastMU, LwSF, SFDN, SSD, and ERM-KTP as baseline methods.~\footnote{We omit ZeroShotMU because it failed to reproduce the forgetting results for CIFAR10, CUB, and STN.
We use the code of https://github.com/ayushkumartarun/zero-shot-unlearning.
}
While some need the training data for forgetting, we confirmed they can work in a realistic time and target class removal.\par
\noindent{\textbf{The comparison results for Forgetting Capability.}}
Table~\ref{ar_results} and~\ref{ef_results} show the comparison results for the MU capability~\footnote{ERM-KTP targets CNNs, so we omit the result for MNIST.}.
We repeat the evaluation three times and take the average of the results within the standard deviation error bars.
The highest values in metrics $A_\mathrm{R}$ and $E_\mathrm{F}$ are shown in bold.
The tables show that our method achieves 100\% $E_\mathrm{F}$ while maintaining high $A_\mathrm{R}$ for all datasets.
We can see that our method is superior to or competitive with the baselines.
Fig.~\ref{time_comparison} shows the comparison results for MU processing time.
These results show that our method works significantly faster than the baselines.
This is because our method can efficiently calculate the FIM for each class with mnemonic code and add effective one-shot perturbation for forgetting, which does not need additional training.
In addition, the MU processing time for ResNet-18 trained on ImageNet is shown in Fig.~\ref{time_comparison}.
This result shows that the MU processing time in FastMU is significantly increased for the large dataset.
On the other hand, our method works quickly for such a dataset.
We also confirm that the other baseline methods do not complete the MU process in a realistic time, which is omitted from Fig.~\ref{time_comparison}.
Detailed scalability evaluations for our method are given in Sec.~\ref{scalability}.

\subsection{Effect of mnemonic code on forgetting}
\label{mncode_effect}
\begin{figure}[tb]
\begin{minipage}{0.48\textwidth}
\begin{center}
\includegraphics[width=\textwidth]{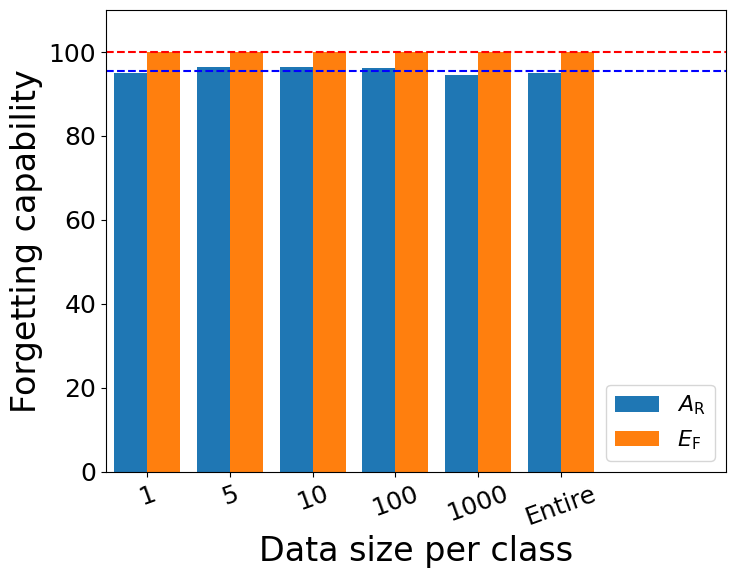}
\end{center}
\subcaption{Forgetting capability on MNIST.}
\end{minipage}
\begin{minipage}{0.48\textwidth}
\begin{center}
\includegraphics[width=\textwidth]{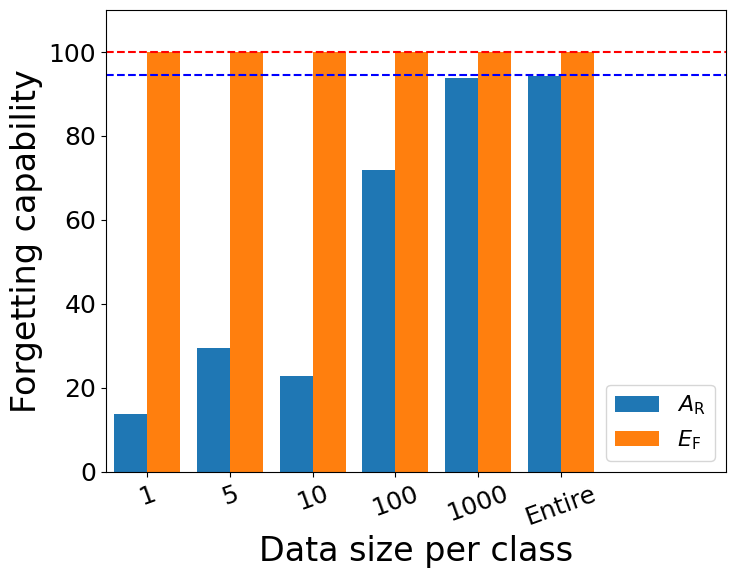}
\end{center}
\subcaption{Forgetting capability on CIFAR10.}
\end{minipage}
\caption{\textbf{The forgetting capability of ours-w-data on MNIST and CIFAR10.}
Our method with mnemonic code is also included for reference in dotted lines.
The blue one shows $A_\mathrm{R}$ and the orange one shows $E_\mathrm{F}$
}
\label{Forgetting_capability_ours_w_data}
\end{figure}

We investigate the effect of mnemonic code on the forgetting capability.
To investigate that, we prepare a baseline method that calculates the perturbation with the training data.
We call this baseline as ours-w-data.
In ours-w-data, we train the model without mnemonic code.
Then, in the forgetting phase, we obtain the FIM with a portion of the training data and add perturbation to the model parameter by Eq.~\ref{addnoise}.
The difference between our method and ours-w-data is whether or not the mnemonic codes are used when training the model and calculating the FIM.
Therefore, we expect that the difference in the forgetting capability comes from the contribution of mnemonic code.
We measure $A_\mathrm{R}$ and $E_\mathrm{F}$ for the two methods and evaluate their forgetting capabilities.
In these experiments, we use MNIST and CIFAR10.
The results are shown in Fig.~\ref{Forgetting_capability_ours_w_data}.
From the results, we can clarify two things in ours-w-data.
First, the forgetting process works effectively when we use all the training data. 
Second, while effective forgetting is possible with a small amount of training data in MNIST, effective forgetting is difficult with a small amount of training data in CIFAR10.
We consider that this difference in datasets is due to the diversity of data contained in each class.
Specifically, a simple dataset such as MNIST has a low diversity of data in each class.
In contrast, a dataset such as CIFAR10 has diverse data in each class, and it will be difficult to represent each class with a small amount of training data.
We consider this difference in datasets led to the difference in forgetting results in ours-w-data.
Figure.~\ref{Forgetting_capability_ours_w_data} also shows the results of our method with mnemonic codes as a reference value.
Surprisingly, our method can effectively forget with only one piece of mnemonic code per class. \par
We then conduct an experiment focusing on the FIM to reveal the cause of the difference in the forgetting capability between our method and ours-w-data shown in Fig.~\ref{Forgetting_capability_ours_w_data}.
We define the FIM calculated using all the training data as the Oracle FIM. 
As seen in Fig.~\ref{Forgetting_capability_ours_w_data}, ours-w-data works well with the Oracle FIM for both MNIST and CIFAR10. 
Also, the Oracle FIM is often used in existing continual learning methods~\citep{kirkpatrick2017overcoming,huszar2018note,ritter2018online}.
We evaluate the approximation error to the Oracle FIM of the FIM calculated using the portion of training data and of the FIM using the mnemonic code.
When calculating the FIM approximation error, we calculate the L2 norm of the FIMs and divide it by the number of elements in the FIM (the number of the model parameters).
\par
\begin{figure}[tb]
\begin{minipage}{0.48\textwidth}
\begin{center}
    \includegraphics[width=\textwidth]{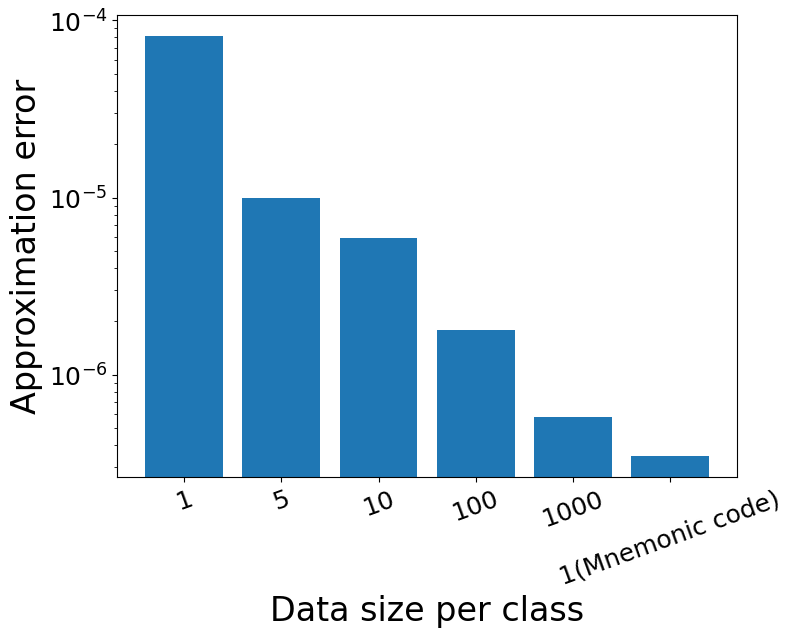}
\end{center}
\subcaption{FIM approximation error on MNIST.}
\end{minipage}
\begin{minipage}{0.48\textwidth}
\begin{center}
\includegraphics[width=\textwidth]{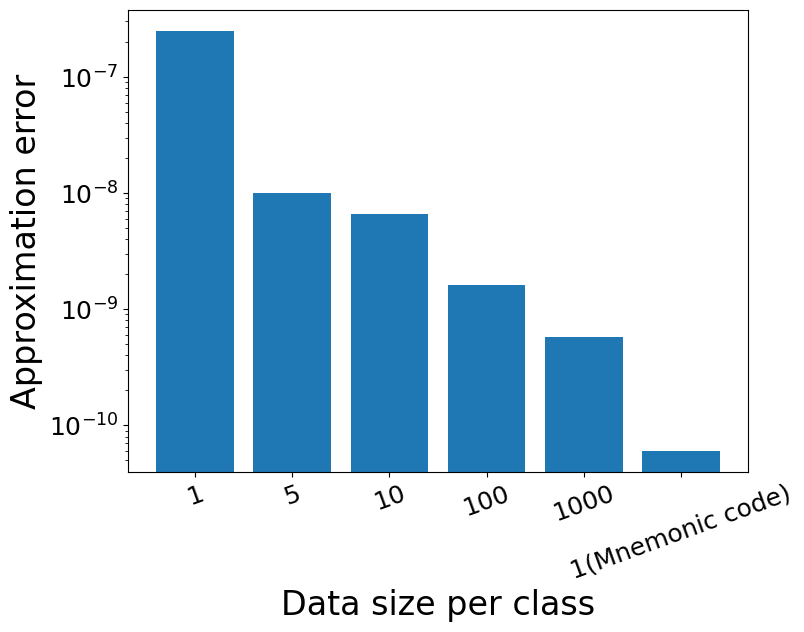}
\end{center}
\subcaption{FIM approximation error on CIFAR10.}
\end{minipage}
\caption{\textbf{The FIM approximation error of ours-w-data on MNIST and CIFAR10.}
Our method with mnemonic code is also included for reference.
}
\label{FIM_approximation_ours_w_data}
\end{figure}

The results are shown in Fig.~\ref{FIM_approximation_ours_w_data}.
The results show that the smaller the number of training data for calculating the FIM, the further away from the Oracle FIM.
Furthermore, we can see that the approximation error for the mnemonic code shown on the right side of the graph is significantly low.
This is a remarkable result since it shows that the FIM obtained with only one mnemonic code is closer to the Oracle FIM than the FIM obtained with 1,000 training data.
The high forgetting performance with mnemonic codes can be explained by the low approximation error of the FIM calculation.
We consider that the low approximation error with mnemonic code is because the mnemonic code is presented to the deep learning model at a high rate of $t_\mathrm{mix}$ during training, while each training data is presented only one time per epoch. 
Furthermore, these results provide validity for replacing the Oracle FIM used in Eq.~\ref{forget_equation} with the FIM calculated with mnemonic code, as in Eq.~\ref{surrogate_loss}. \par

\begin{table}[tb]
\small
    \centering
    \caption{\textbf{Forgetting for ImageNet dataset.}
    We perform fine-tuning using mnemonic code on the pre-trained model and forget with our method.
    We show the forgetting capability and processing time for the pre-trained, fine-tuned, and forgotten models, respectively.
    }
    \begin{tabular}{llccc}
        \hline
        Architecture&& $A_\mathrm{R} \uparrow$ & $E_\mathrm{F} \uparrow$ & Time [s] $\downarrow$ \\
        \hline
        \multirow{3}{*}{ResNet-18} & Pretrained & 69.8 & 12.0 & - \\
        &Fine-tuned & 67.5 & 12.0 & 882 \\
        &After MU & 67.5 & 100 & 8.66 \\
        \hline
        \multirow{3}{*}{ResNeXt-50} & Pretrained & 77.4 & 6.0 & - \\
        &Fine-tuned & 75.9 & 12.0 & 6923 \\
        &After MU & 75.9 & 100 & 21.0 \\
        \hline
        \multirow{3}{*}{Swin-Transformer} & Pretrained & 80.9 & 4.0 & - \\
        &Fine-tuned & 78.8 & 4.0 & 8488 \\
        &After MU & 75.3 & 92.0 & 28.6 \\
        \hline
    \end{tabular}
    \label{imagenet_results}
\end{table}

\subsection{Scalability}
\label{scalability}
As shown in Fig.~\ref{time_comparison}, our method works much faster than the baseline methods.
This section demonstrates that our lightweight method is scalable to large practical datasets and sophisticated models.
In the experiments, we prepare the pre-trained model and fine-tune it for a few steps using mnemonic codes.
Then, we apply our method to the fine-tuned model.
We evaluate the forgetting capability with $A_{\mathrm R}$ and $E_\mathrm{F}$, and the processing time for fine-tuning and MU processing.
The dataset is ImageNet, consisting of 1,000 classes, and the architecture is ResNet-18, ResNeXt-50, and Swin-Transformer.
The number of fine-tuning steps is 2,000 for ResNet-18 and 10,000 for ResNeXt-50 and Swin-Transformer. \par
Table.~\ref{imagenet_results} shows that our method works within one minute and improves $E_\mathrm{F}$ without significantly reducing $A_\mathrm{R}$. 
It was also found that fine-tuning with mnemonic codes takes several hours.
Therefore, if we fine-tune the pre-trained model with mnemonic codes in advance, we can quickly perform the MU process when an unlearning request arises. 
The results also show that our method can achieve effective forgetting for pre-trained models.
Note that Table.~\ref{imagenet_results} shows that test accuracy slightly degrades due to fine-tuning using mnemonic codes.
We consider this because the number of classes in ImageNet is large, and the patterns of randomly generated mnemonic codes are insufficient.
Future work will include obtaining mnemonic codes that maintain accuracy, even for large class datasets.
\par

\vspace{0.05cm}
\noindent{\textbf{Limitation.}}
We have found that multi-class forgetting is difficult with our method.
Even in existing methods, multi-class forgetting methods are mainly based on additional training~\citep{tarun2023fast,chundawat2023zero,lin2023erm}, and multiple classes are difficult to forget with one-shot perturbation.
Nevertheless, our method is valuable in that its overwhelmingly faster processing speed and its high scalability will enable us to respond rapidly to unlearning requests, as described in Sec.~\ref{intro}.
Also, we did not fully evaluate our method concerning privacy. 
However, we evaluate the robustness against membership inference attacks and backdoor attacks in the supplemental material.

\section{Future work and Conclusion}
This paper proposes a one-shot MU method that achieves forgetting by adding perturbation to the model parameter.
Mnemonic code is used to reduce the processing time and the computational costs. 
We experimentally showed that our method is lightweight and effective.
Also, we experimentally demonstrated the effectiveness of mnemonic code. 
Furthermore, our method is scalable to more practical datasets and sophisticated architectures.
In future work, we will seek mnemonic codes that do not degrade accuracy even for large class datasets.
Also, we will seek a method that can forget multi-class effectively.
In addition, generative models such as text-to-image models and large-language-models are rapidly advancing as a new application, and the issue of copyright of the data contained in the training data is surfacing~\citep{carlini2021extracting,carlini2023extracting}. 
Lightweight MU methods that can be applied to such sophisticated AI models should be required in the near future.
We believe this paper will contribute to the practical MU research and new possibilities of lightweight MU.

\bibliography{acml24}

\appendix

\section{Experimental Setting}
\label{experimental_setting}
\begin{figure}[tb]
\small
\begin{minipage}{0.48\linewidth}
    \centering
    \includegraphics[width=\linewidth]{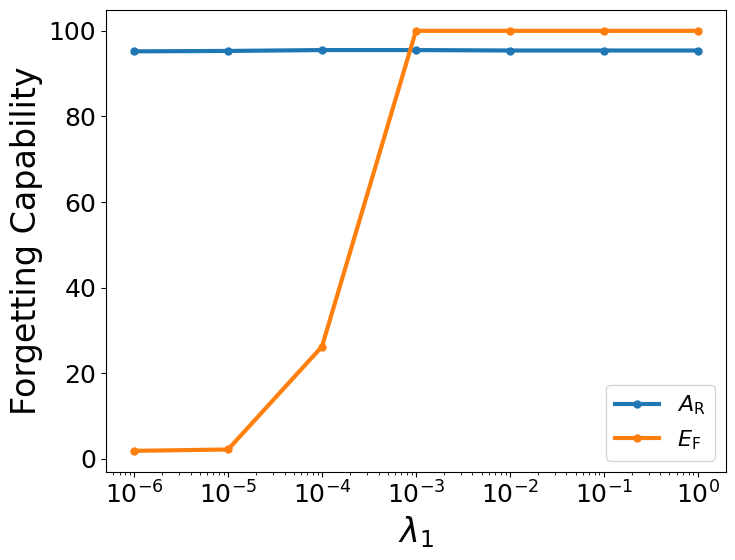}
    \subcaption{$\lambda_1$ search.}
\end{minipage}
\begin{minipage}{0.48\linewidth}
    \centering
    \includegraphics[width=\linewidth]{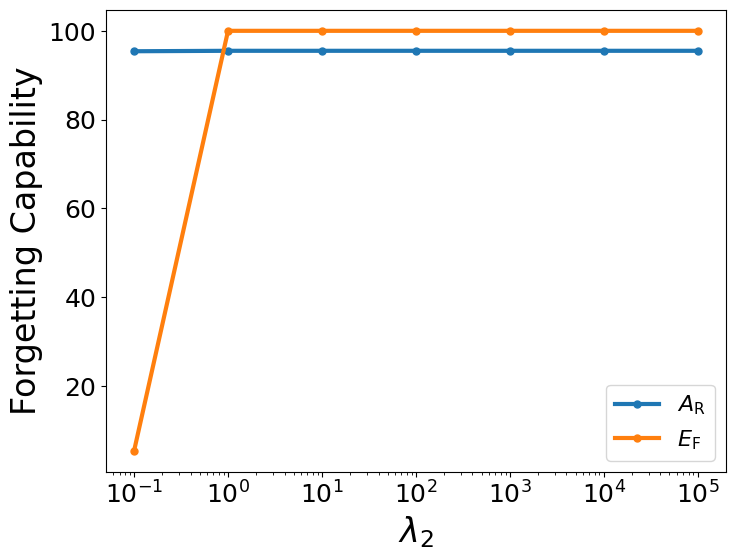}
    \subcaption{$\lambda_2$ search.}
\end{minipage}
\caption{\textbf{Hyperparameter search on MNIST.}}
\label{parasearch_mnist}
\end{figure}

\begin{figure}[tb]
\small
\begin{minipage}{0.48\linewidth}
    \centering
    \includegraphics[width=\linewidth]{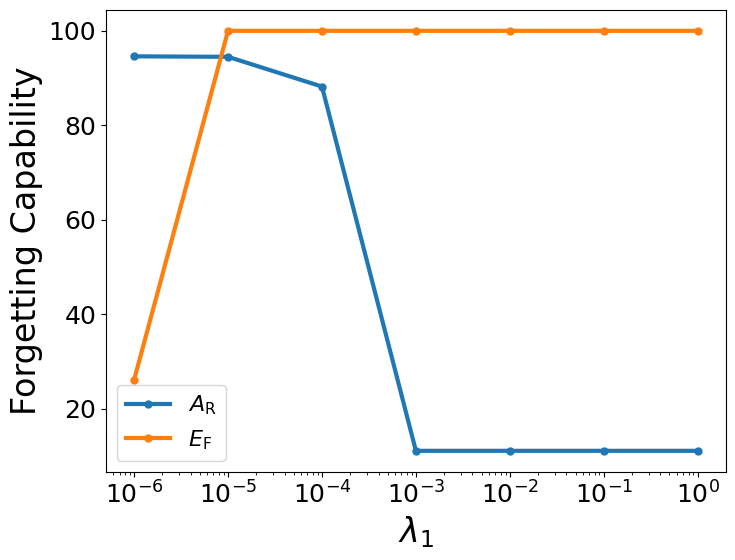}
    \subcaption{$\lambda_1$ search.}
\end{minipage}
\begin{minipage}{0.48\linewidth}
    \centering
    \includegraphics[width=\linewidth]{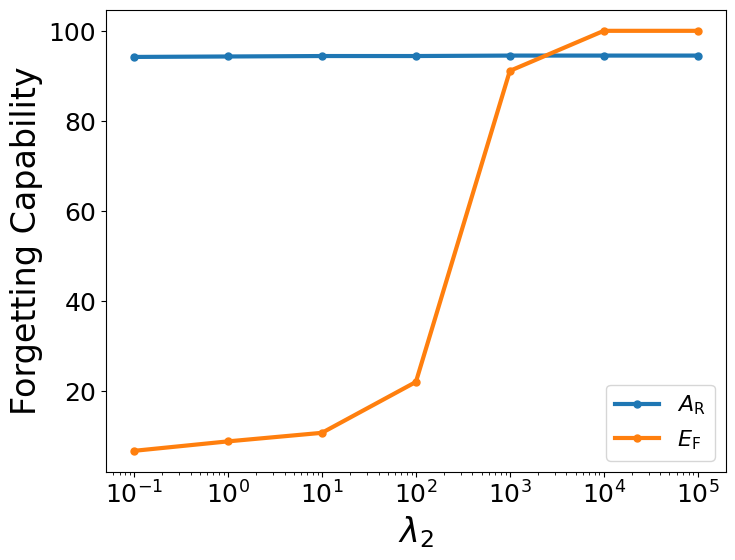}
    \subcaption{$\lambda_2$ search.}
\end{minipage}
\caption{\textbf{Hyperparameter search on CIFAR10.}}
\label{parasearch_cifar}
\end{figure}

\begin{figure}[tb]
\small
\begin{minipage}{0.48\linewidth}
    \centering
    \includegraphics[width=\linewidth]{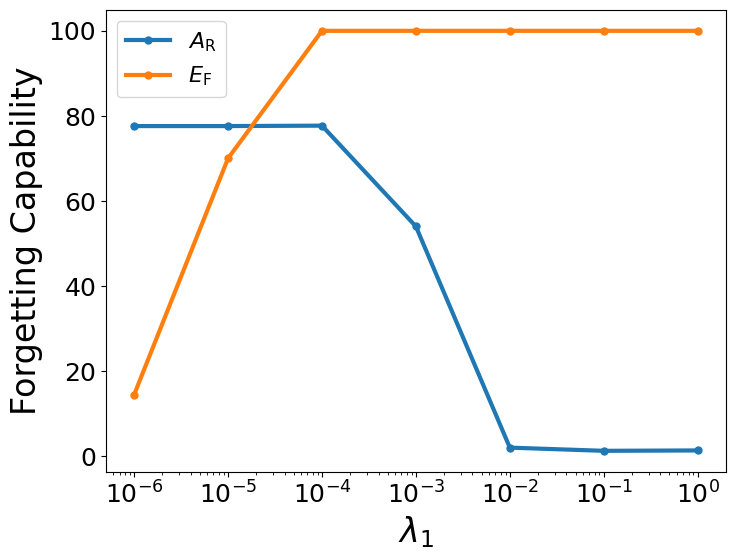}
    \subcaption{$\lambda_1$ search.}
\end{minipage}
\begin{minipage}{0.48\linewidth}
    \centering
    \includegraphics[width=\linewidth]{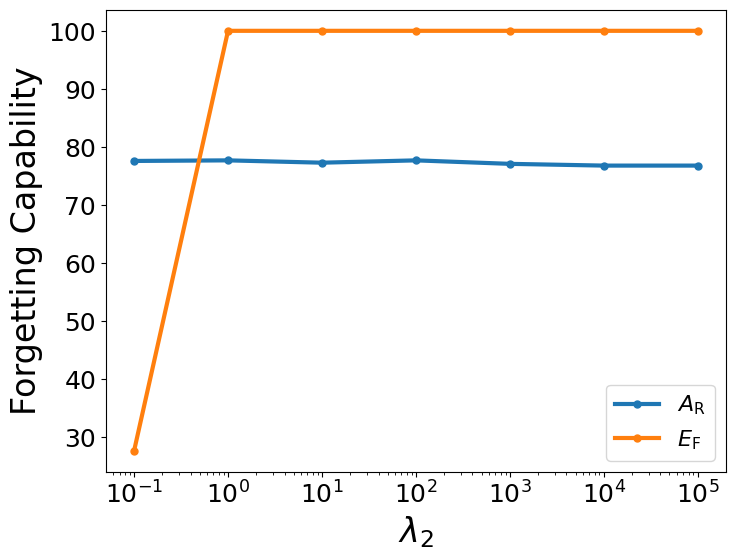}
    \subcaption{$\lambda_2$ search.}
\end{minipage}
\caption{\textbf{Hyperparameter search on CUB.}}
\label{parasearch_cub}
\end{figure}

\begin{figure}[tb]
\small
\begin{minipage}{0.48\linewidth}
    \centering
    \includegraphics[width=\linewidth]{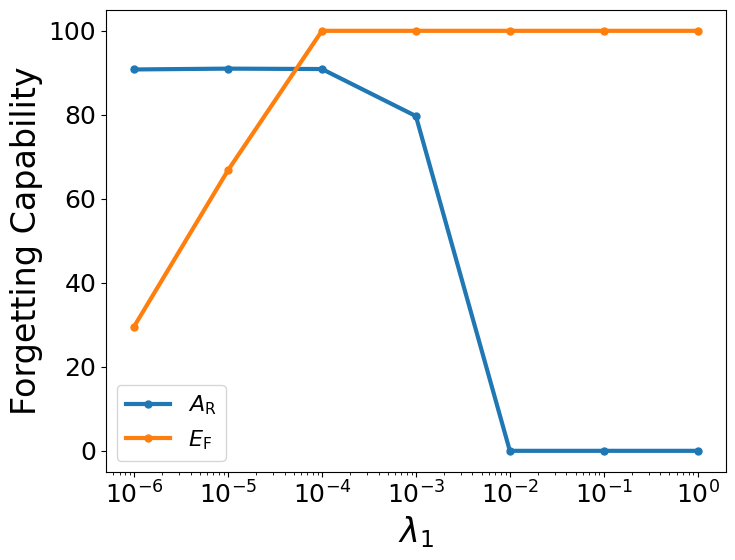}
    \subcaption{$\lambda_1$ search.}
\end{minipage}
\begin{minipage}{0.48\linewidth}
    \centering
    \includegraphics[width=\linewidth]{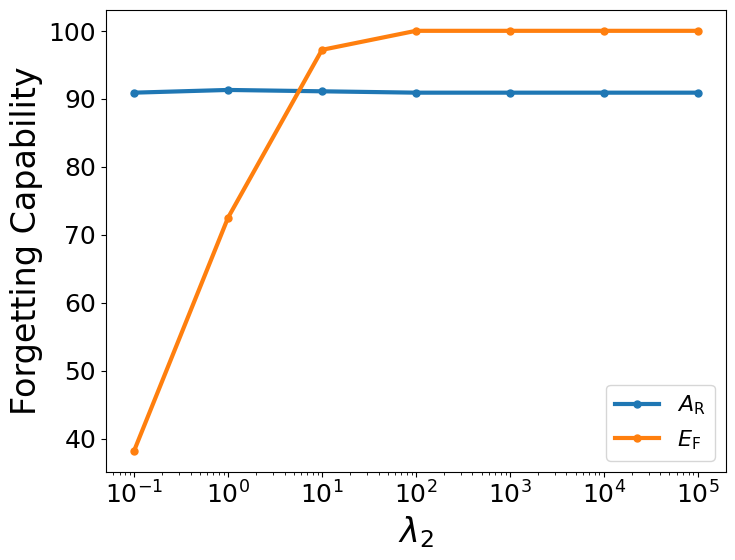}
    \subcaption{$\lambda_2$ search.}
\end{minipage}
\caption{\textbf{Hyperparameter search on STN.}}
\label{parasearch_stn}
\end{figure}

We describe the experimental settings.
We use the SGD optimizer for all datasets and train each model for 200 epochs.
The learning rate is set to 0.01, and weight decay is set to $5 \times 10^{-4}$.
We use a cosine scheduler for CIFAR10, CUB, and STN.
The batch size is 128 for MNIST and CIFAR10 and 32 for CUB and STN.
In fine-tuning with mnemonic code for ImageNet, the learning rate is set to 0.001, weight decay is $5 \times 10^{-4}$, and the batch size is 64.
For the hyperparameter set, we search $\lambda_1=[10^{-6}, 10^{-5}, \cdots, 1.0]$ and $\lambda_2=[10^{-1}, 1.0, \cdots, 10^5]$ and select the combination that maximizes the sum of $A_\mathrm{R}$ and $E_\mathrm{F}$.
We use $(\lambda_1, \lambda_2)=(10^{-3}, 10.0)$ for MNIST, $(\lambda_1, \lambda_2)=(10^{-5}, 10^4)$ for CIFAR10, $(\lambda_1, \lambda_2)=(10^{-4}, 1.0)$ for CUB, and $(\lambda_1, \lambda_2)=(10^{-4}, 100.0)$ for STN.
Here, we show the results of the hyperparameter search.
We evaluate the forgetting capability by fixing one of $\lambda_1$ and $\lambda_2$ to the above values while changing the other.
The results are shown on Fig.~\ref{parasearch_mnist},~\ref{parasearch_cifar},~\ref{parasearch_cub}, and~\ref{parasearch_stn}.
These results show a tradeoff between $E_\mathrm{F}$ improving while $A_\mathrm{R}$ decreasing as $\lambda_1$ increased.
On the other hand, as $\lambda_2$ is increased, $E_\mathrm{F}$ improves while $A_\mathrm{R}$ does not significantly decrease.

\begin{figure}[tb]
\small
    \begin{tabular}{cc}
      \begin{minipage}[t]{0.48\hsize}
        \centering
        \includegraphics[width=\linewidth]{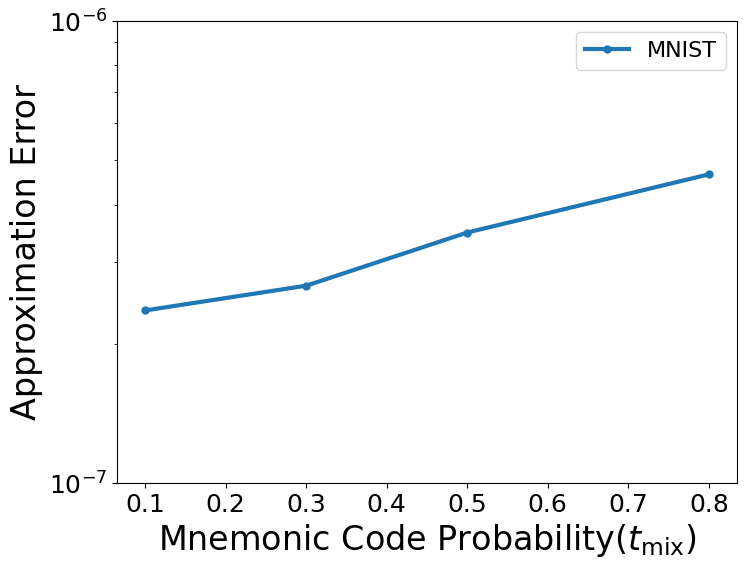}
        \subcaption{Result on MNIST.}
      \end{minipage} &
      \begin{minipage}[t]{0.48\hsize}
        \centering
        \includegraphics[width=\linewidth]{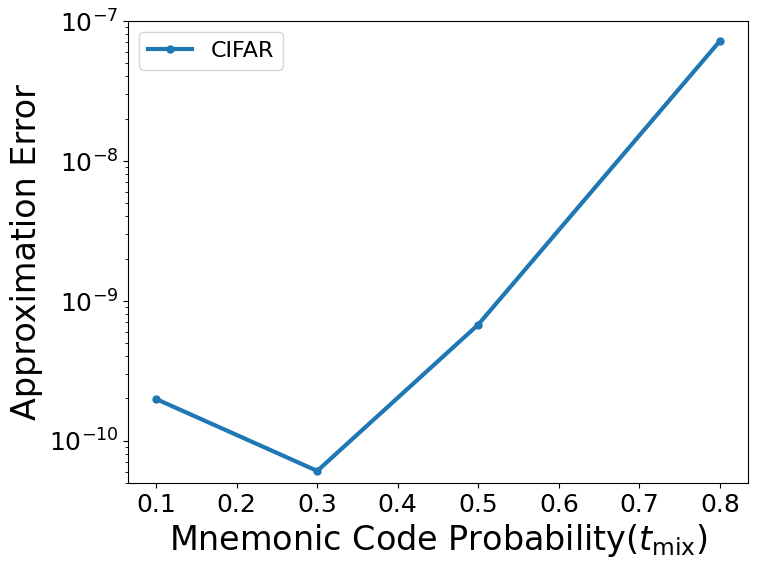}
        \subcaption{Result on CIFAR10.}
      \end{minipage} \\
   
      \begin{minipage}[t]{0.48\hsize}
        \centering
        \includegraphics[width=\linewidth]{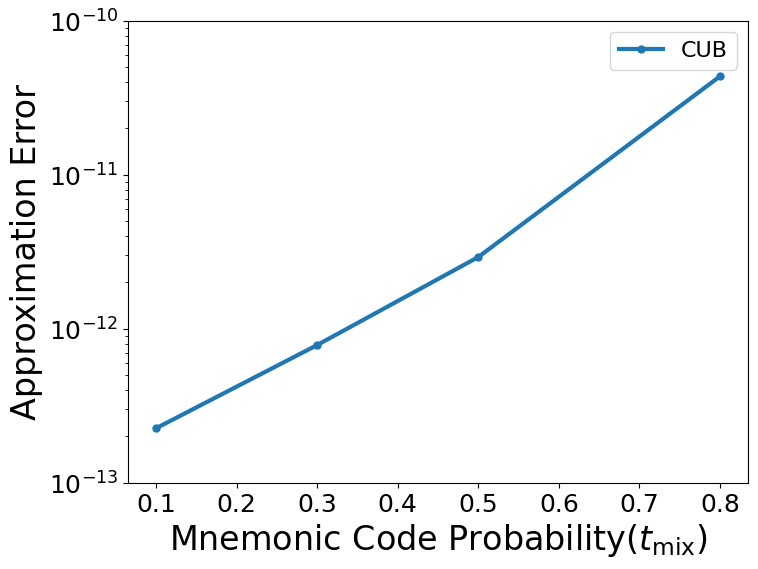}
        \subcaption{Result on CUB.}
      \end{minipage} &
      \begin{minipage}[t]{0.48\hsize}
        \centering
        \includegraphics[width=\linewidth]{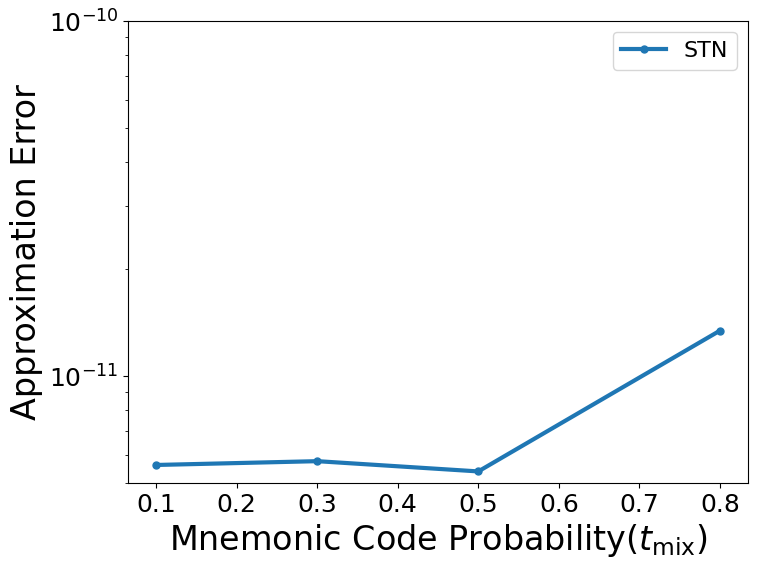}
        \subcaption{Result on STN.}
      \end{minipage} 
    \end{tabular}
    \caption{\textbf{FIM approximation with various $t_\mathrm{mix}$.}
    }
    \label{t_mix_FIM}
\end{figure}

Then, we consider the setting of $t_\mathrm{mix}$, the probability of replacing the training data with mnemonic codes.
We search $t_\mathrm{mix}=[0.1, 0.3, 0.5, 0.8]$ and select the best value that maximizes the forgetting capability.
The results are shown in Table~\ref{t_mix_search}.
The results show that $E_\mathrm{F}$ achieves $100\%$ for all cases.
When $t_\mathrm{mix}=0.1$, the sum of $E_\mathrm{F}$ and $A_\mathrm{R}$ is maximum in MNIST, CUB, and STN, and when $t_\mathrm{mix}=0.3$, the sum of $E_\mathrm{F}$ and $A_\mathrm{R}$ is maximum in CIFAR10. 
Furthermore, we evaluate the approximation errors to the Oracle FIM of the FIM obtained with the mnemonic codes when changing $t_\mathrm{mix}$, shown in Fig.~\ref{t_mix_FIM}.
The results show that the approximation error is minimum when $t_\mathrm{mix}=0.1$ in MNIST and CUB, and the error is minimum when $t_\mathrm{mix}=0.3$ in CIFAR10.
In STN, the approximation errors are almost the same when $t_\mathrm{mix}$ is 0.1, 0.3 and 0.5.
These results are consistent with the evaluation results of the forgetting capability when $t_\mathrm{mix}$ is changed. 
It also shows that there is no significant difference in forgetting capability when $t_\mathrm{mix}$ is 0.1 and 0.3.
Therefore, these results lead to the conclusion that setting $t_\mathrm{mix}$ below 0.3 is reasonable in our method. \par
When conducting our experiments in the main paper, we performed the hyperparameter search described above to set up $\lambda_1$, $\lambda_2$, and $t_\mathrm{mix}$.
It should be noted that because our method works significantly fast, we can perform such hyperparameter searches quickly.

\begin{table}[tb]
\small
\tabcolsep=5pt
    \centering
    \caption{
    \textbf{Forgetting capability on various $t_\mathrm{mix}$.}
    The table shows the forgetting capability of our method with different $t_\mathrm{mix}$. 
    It shows that the forgetting capability is high at $t_\mathrm{mix}$ of 0.1 or 0.3.
    It also shows that there is no significant difference in forgetting capability when $t_\mathrm{mix}$ is below 0.3.
    }
    \begin{tabular}{lcccccccc}
        \hline
        & \multicolumn{2}{c}{MNIST} & \multicolumn{2}{c}{CIFAR10} & \multicolumn{2}{c}{CUB} & \multicolumn{2}{c}{STN} \\
        $t_\mathrm{mix}$ & $A_\mathrm{R} \uparrow$ & $E_\mathrm{F} \uparrow$ & $A_\mathrm{R} \uparrow$ & $E_\mathrm{F} \uparrow$ & $A_\mathrm{R} \uparrow$ & $E_\mathrm{F} \uparrow$ & $A_\mathrm{R} \uparrow$ & $E_\mathrm{F} \uparrow$ \\
        \hline
        0.1 & 95.9 & 100 & 94.2 & 100 & 79.3 & 100 & 91.7 & 100 \\
        0.3 & 95.4 & 100 & 94.4 & 100 & 77.5 & 100 & 89.3 & 100 \\
        0.5 & 94.2 & 100 & 94.3 & 100 & 75.3 & 100 & 90.3 & 100 \\
        0.8 & 91.1 & 100 & 68.2 & 100 & 68.2 & 100 & 90.2 & 100 \\
        \hline
    \end{tabular}
    \label{t_mix_search}
\end{table}

\section{Validity of Laplace's approximation}
\label{validity}

We investigate the validity of the assumptions of the Laplace approximation, $\nabla \mathcal{L}_{\mathcal{C}_{\mathrm{F}}}(\bm{w}^*)=0$ and $\nabla \mathcal{L}_{\mathcal{C}_{\mathrm{R}}}(\bm{w}^*)=0$, applied in Eq.~\ref{forget_equation}. 
In general, $\nabla \mathcal{L}(\bm{w}^*)=0$ is valid for trained models.
Thus, if $|\nabla \mathcal{L}_{\mathcal{C}_{\mathrm{F}}}(\bm{w}^*)|$ and $|\nabla \mathcal{L}_{\mathcal{C}_{\mathrm{R}}}(\bm{w}^*)|$ are close enough to $|\nabla \mathcal{L}(\bm{w}^*)|$, then we can verify the assumptions of the Laplace approximation.
In the experiments, we train ResNet-18 on CIFAR10 with mnemonic codes and investigate the magnitude of loss gradient, $|\nabla \mathcal{L}(\bm{w}^*)|$, $|\nabla \mathcal{L}_{\mathcal{C}_{\mathrm{F}}}(\bm{w}^*)|$, and $|\nabla \mathcal{L}_{\mathcal{C}_{\mathrm{R}}}(\bm{w}^*)|$ in each layer.
The results are shown in Table.~\ref{laplace_valid}.
These results indicate that $|\nabla \mathcal{L}_{\mathcal{C}_{\mathrm{F}}}(\bm{w}^*)|$ and $|\nabla \mathcal{L}_{\mathcal{C}_{\mathrm{R}}}(\bm{w}^*)|$ are extremely small, about $10^{-5}$.
In addition, $|\nabla \mathcal{L}_{\mathcal{C}_{\mathrm{F}}}(\bm{w}^*)|$ and $|\nabla \mathcal{L}_{\mathcal{C}_{\mathrm{R}}}(\bm{w}^*)|$ are of almost same order of $|\nabla \mathcal{L}(\bm{w}^*)|$ in all layers.
From these perspectives, we consider that the assumption $\nabla \mathcal{L}_{\mathcal{C}_{\mathrm{F}}}(\bm{w}^*)=0$ and $\nabla \mathcal{L}_{\mathcal{C}_{\mathrm{R}}}(\bm{w}^*)=0$ are valid.

\begin{table}[tb]
\small
    \centering
    \caption{
    \textbf{
    The loss gradient magnitude in each layer.
    }
    }
    \begin{tabular}{lccccc}
        \hline
        Loss gradient magnitude & layer1 & layer2 & layer3 & layer4 & layer5 \\
        \hline
        $|\nabla \mathcal{L}(\bm{w}^*)|$ & $6.9\times 10^{-6}$ & $5.6\times 10^{-6}$ & $7.4\times 10^{-6}$ & $9.2\times 10^{-6}$ & $1.6\times 10^{-5}$ \\
        $|\nabla \mathcal{L}_{\mathcal{C}_{\mathrm{F}}}(\bm{w}^*)|$ & $3.6\times 10^{-5}$ & $3.0\times 10^{-5}$ & $1.8\times 10^{-5}$ & $1.2\times 10^{-5}$ & $1.4\times 10^{-5}$ \\
        $|\nabla \mathcal{L}_{\mathcal{C}_{\mathrm{R}}}(\bm{w}^*)|$ & $1.3\times 10^{-5}$ & $9.9\times 10^{-6}$ & $8.1\times 10^{-6}$ & $7.6\times 10^{-6}$ & $9.4\times 10^{-6}$ \\
        \hline
    \end{tabular}
    \label{laplace_valid}
\end{table}

\section{Additional experiments}

\subsection{Evaluation on membership inference attack}
\label{mia_evaluation}
\begin{figure}[tb]
    \begin{minipage}{0.48\textwidth}
    \begin{center}
    \includegraphics[width=\textwidth]{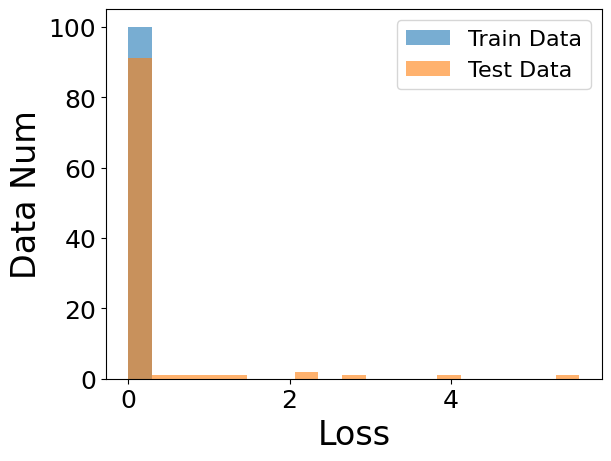}
    \end{center}
    \subcaption{Before forgetting.}
    \label{before_forget}
    \end{minipage}
    \begin{minipage}{0.48\textwidth}
    \begin{center}
        \includegraphics[width=\textwidth]{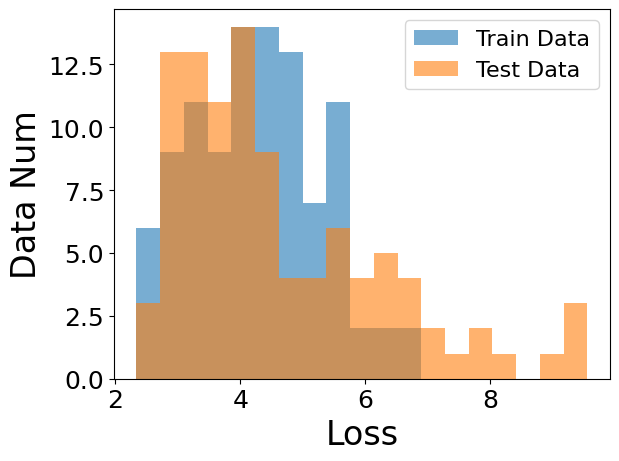}
    \end{center}
    \subcaption{After forgetting.}
    \label{after_forget}
    \end{minipage}
\caption{\textbf{Loss distribution on the models before and after our forgetting.}}
\end{figure}

Membership inference attack (MIA) is one of the AI security attacks that are of concern from a privacy perspective~\citep{shokri2017membership}.
In this attack, individual data is inferred to be included in the training data of the trained model.
Typically, the attacker inputs data into the model and observes the output loss. 
Then, the value of the loss is used to determine if the input data is included in the training data.
Here, we investigate whether our method is robust to MIA.
Specifically, we apply our method to the trained model and compare the loss distribution of training data in $\mathcal{C}_\mathrm{F}$ with the loss distribution of test data in $\mathcal{C}_\mathrm{F}$. 
We train ResNet-18 on CIFAR10 with mnemonic codes and apply our forgetting method to the model.
The settings of the hyperparameters are consistent with the main paper.
The results are shown in Fig.~\ref{before_forget} and~\ref{after_forget}.
Figure.~\ref{before_forget} shows the result of the loss distribution of the trained model, and Fig.~\ref{after_forget} shows the result of the forgotten model.
These results show that the loss distribution of the training data and the loss distribution of the test data overlap significantly in the forgotten model and are robust to MIA.
In fact, we confirm that the AUC for the naive MIA using these loss distributions is 0.49, describing that it is difficult to separate training data from test data on the forgotten model. \par
\begin{figure}[tb]
    \begin{minipage}{0.48\textwidth}
    \begin{center}
    \includegraphics[width=\textwidth]{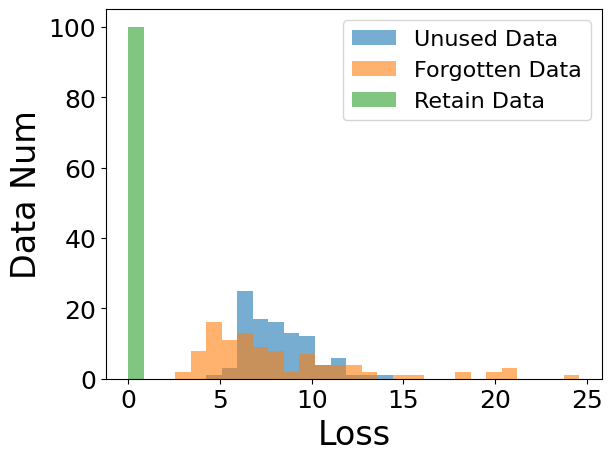}
    \end{center}
    \subcaption{The model forgotten with our method.}
    \label{class_mia_forget}
    \end{minipage}
    \begin{minipage}{0.48\textwidth}
    \begin{center}
        \includegraphics[width=\textwidth]{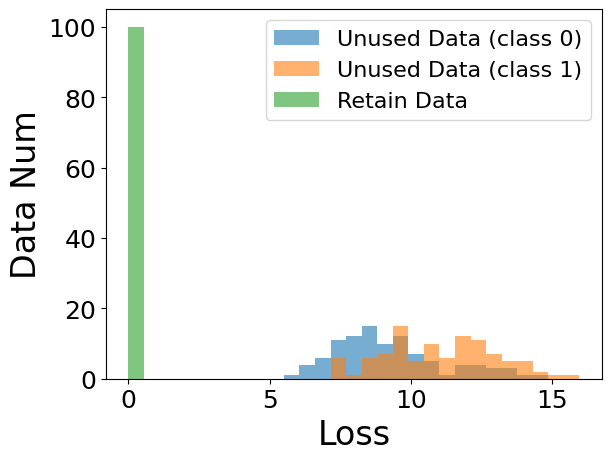}
    \end{center}
    \subcaption{The model trained without class 0 and 1.}
    \label{class_mia_unused}
    \end{minipage}
\caption{\textbf{Loss distributions on some AI models.}}
\end{figure}

We then investigate whether the loss distribution of the forgotten class is sufficiently close to the loss distribution of the class that was not used for training.
If there is a large discrepancy between the two loss distributions, we consider that an attacker may be able to identify the forgotten class from the loss distributions.
In this experiment, we train ResNet-18 without class 0.
Then, we forget the model about class 1 and investigate the loss distribution of class 0 and class 1.
Therefore, the data in class 0 are unused, and the data in class 1 are forgotten.
The result is shown in Fig.~\ref{class_mia_forget}.
This result shows the loss distribution of the unused, forgotten, and remaining data.
The result shows that the loss distributions of unused and forgotten data roughly overlap but diverge slightly. \par
We suspect that the class difference between Class 0 and Class 1 may contribute to the discrepancy between the loss distributions, and we will conduct further experiments.
We proceed to conduct an experiment to see how the class difference affects the difference in the loss distributions.
We train ResNet-18 without class 0 and class 1 and investigate the loss distributions of the two classes.
The result is shown in Fig.~\ref{class_mia_unused}.
This result shows that the loss distribution of class 0 and that of class 1 diverge slightly.
Therefore, we believe the above hypothesis is correct. 
From these results, we show that our method does not make the model vulnerable to MIA.

\subsection{The meaning of \texorpdfstring{$E_\mathrm{F}=100$}{E\mathrm{F}=100} in class forgetting}
We confirmed by experiments in the main paper that our method can achieve $E_\mathrm{F}=100$.
In this section, we will see how an AI model with $E_\mathrm{F}=100$ behaves concerning MNIST and CIFAR10.
In the experiment, we prepared a model that forget class 0 with our method and an Oracle model trained without the data of class 0.
We then input test data to those AI models and examine the rate at which they output class 0. 
As a result of the experiment, we confirmed that these AI models do not output class 0 for any given test data.
Therefore, from the results, we can confirm that both forgotten models by our method and the Oracle models have similar behavior in that they do not output the target class.

\subsection{Evaluation on backdoor attack}
\label{evaluate_backdoor}
We investigate whether our method makes the model vulnerable to backdoor attacks.
Backdoor attacks indicate that an attacker has planted a backdoor in the AI model~\citep{li2022backdoor}.
The attack technique is such that the model makes accurate predictions for general input data and incorrect predictions when the attacker uses a specific pattern (backdoor trigger).
In our method, when training the model, we present the mnemonic codes to the model at a high rate of $t_\mathrm{mix}$.
We investigate whether the mnemonic code in our method can act as a backdoor trigger.
In the experiment, we train ResNet-18 on CIFAR10 using mnemonic codes with $t_\mathrm{mix}=0.3$.
Then, we evaluate test accuracy using test images mixed with class 0 mnemonic code.
If our method makes the model vulnerable to backdoor attacks, the model's output on the test data with the mnemonic code will be drawn to class 0, and we should see accuracy degradation.
The mixing ratio of mnemonic codes is set to $\left[0\%, 10\%, 30\%, 50\%, 80\%, 100\% \right]$.
When the mixing ratio is set to 0\%, the accuracy is for test images without the mnemonic code, and if 100\%, the test image is a complete class 0 mnemonic code. \par
\begin{figure}[tb]
\centering
\includegraphics[width=0.6\textwidth]{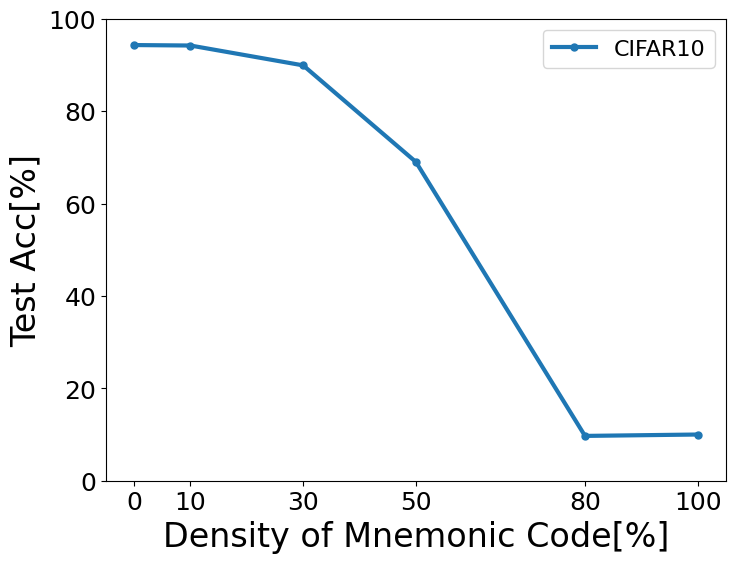}
\caption{
\textbf{Test accuracy for images mixed with mnemonic code.}
}
\label{backdoor_acc}
\end{figure}
\begin{figure}[tb]
\small
    \begin{tabular}{ccc}
      \begin{minipage}[t]{0.32\hsize}
        \centering
        \includegraphics[width=0.7\linewidth]{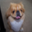}
        \subcaption{0\% code.}
      \end{minipage} &
      \begin{minipage}[t]{0.32\hsize}
        \centering
        \includegraphics[width=0.7\linewidth]{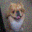}
        \subcaption{10\% code.}
      \end{minipage} &
      \begin{minipage}[t]{0.32\hsize}
        \centering
        \includegraphics[width=0.7\linewidth]{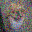}
        \subcaption{30\% code.}
      \end{minipage}
      \\

      \begin{minipage}[t]{0.32\hsize}
        \centering
        \includegraphics[width=0.7\linewidth]{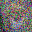}
        \subcaption{50\% code.}
      \end{minipage} &
      \begin{minipage}[t]{0.32\hsize}
        \centering
        \includegraphics[width=0.7\linewidth]{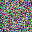}
        \subcaption{80\% code.}
      \end{minipage} &
      \begin{minipage}[t]{0.32\hsize}
        \centering
        \includegraphics[width=0.7\linewidth]{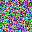}
        \subcaption{100\% code.}
      \end{minipage}
      \\
    \end{tabular}
    \caption{\textbf{Images with mnemonic code at various densities.}
    }
    \label{backdoor_images}
\end{figure}

The evaluation results are shown in Fig.~\ref{backdoor_acc}.
These results show that the test accuracy is not significantly degraded for test images mixed with mnemonic code at a ratio of 10\%.
In fact, the test accuracy is 94.3\% when the ratio of mnemonic code is 0\% and 94.2\% when the ratio is 10\%.
In addition, test images with the mnemonic code mixed at different ratios are shown in Fig.~\ref{backdoor_images}.
These images confirm that images mixed with the mnemonic code of 30\% or higher ratio are difficult for the human eye to discriminate.
These results confirm that mnemonic code does not work as a backdoor trigger in our method.

\subsection{Evaluation on face information}
Here, we conduct the evaluation experiments on face information.
We use Labeled Faces in the Wild (LFW)~\citep{LFWTech}, a dataset used for benchmarking unconstrained face verification performance. 
We created a 100-class discriminative model by training ResNet-18 on this dataset and performed class 0 forgetting with our method and investigate the forgetting capability before and after our forgetting process. 
$t_\mathrm{mix}$ is set to 0.1.
The results are shown in Table~\ref{face_evaluation}.
It shows that our method achieves $100\%$ $E_\mathrm{F}$ and does not damage $A_\mathrm{R}$ significantly.
It demonstrates our method is effective for face information.

\begin{table}[tb]
\small
    \centering
    \caption{\textbf{Forgetting capability for face information.}
    }
    \begin{tabular}{lcc}
        \hline
        Forgetting capability & $A_\mathrm{R} \uparrow$ & $E_\mathrm{F} \uparrow$ \\
        \hline
        Before Forget & 76.3 & 22.2 \\
        After Forget & 76.2 & 100 \\
        \hline
    \end{tabular}
    \label{face_evaluation}
\end{table}

\subsection{Forgetting other classes}
\label{forget_otherclass}
We evaluate the forgetting capability of our method on classes other than 0.
As with the main paper, we use a simple, fully connected model for MNIST and ResNet-18 for CIFAR10, CUB, and STN.
We evaluate $A_\mathrm{R}$ and $E_\mathrm{F}$ for forgetting capability.
In this experiment, we forget classes 1 to 9.
The results are shown in Table~\ref{otherclass_unlearn}. \par
Table~\ref{otherclass_unlearn} shows almost no difference in forgetting capability among classes.
Thus, it indicates that our method can calculate the effective perturbation for forgetting and work effectively regardless of which class of the forgetting target.

\begin{table}[tb]
    \centering
    \caption{
    \textbf{Forgetting other classes.}
    The table shows the results of applying our method to classes 1 to 9.
    }
    \begin{tabular}{lcccccccc}
        \hline
        & \multicolumn{2}{c}{MNIST} & \multicolumn{2}{c}{CIFAR10} & \multicolumn{2}{c}{CUB} & \multicolumn{2}{c}{STN} \\
        Forget class & $A_\mathrm{R} \uparrow$ & $E_\mathrm{F} \uparrow$ & $A_\mathrm{R} \uparrow$ & $E_\mathrm{F} \uparrow$ & $A_\mathrm{R} \uparrow$ & $E_\mathrm{F} \uparrow$ & $A_\mathrm{R} \uparrow$ & $E_\mathrm{F} \uparrow$ \\
        \hline
        1 & 95.4 & 100 & 93.5 & 100 & 78.0 & 100 & 90.7 & 100 \\
        2 & 95.4 & 100 & 94.4 & 100 & 78.9 & 100 & 90.6 & 100 \\
        3 & 95.6 & 100 & 95.4 & 100 & 77.3 & 100 & 90.3 & 100 \\
        4 & 96.0 & 100 & 93.9 & 100 & 77.6 & 100 & 89.5 & 100 \\
        5 & 96.0 & 100 & 95.0 & 99.8 & 78.1 & 100 & 89.7 & 100 \\
        6 & 95.4 & 100 & 94.3 & 100 & 77.9 & 100 & 90.1 & 100 \\
        7 & 96.1 & 100 & 91.2 & 100 & 78.4 & 100 & 90.7 & 100 \\
        8 & 95.6 & 100 & 93.6 & 100 & 79.2 & 100 & 89.8 & 100 \\
        9 & 96.3 & 100 & 93.6 & 100 & 77.7 & 100 & 90.9 & 100 \\
        \hline
    \end{tabular}
    \label{otherclass_unlearn}
\end{table}

\subsection{Evaluation on the model with CLIP}
\label{evaluate_clip}
We evaluate our method to the model with CLIP-style loss~\citep{radford2021learning}.
CLIP-style loss uses CLIP text encodings of the class names as labels.
In such models, classes are represented in the text encoding space, so the number of classes is virtually infinite.
Therefore, class forgetting cannot be simplified for such models by cutting the edge in the final linear layer, because they have an infinite number of classes and it is difficult to find the edge corresponding to the forgetting target class.
In this experiment, we train ResNet18 on CIFAR10 using CLIP-style loss instead of cross-entropy loss.
We then evaluate the forgetting capability of our method by forgetting the class ``airplane''.
The result is shown in Table.~\ref{clip_evaluation}.
It shows that our method achieves 100\% $E_\mathrm{F}$ and does not damage $A_\mathrm{R}$.
It demonstrates our method is effective for the model with CLIP-style loss.
Furthermore, this result also shows that our method is not a process that only cuts off the edge corresponding to the forgetting target class in the final linear layer.

\begin{table}[tb]
\small
    \centering
    \caption{\textbf{Forgetting capability to the model with CLIP-style loss.}
    }
    \begin{tabular}{lcc}
        \hline
        Forgetting capability & $A_\mathrm{R} \uparrow$ & $E_\mathrm{F} \uparrow$ \\
        \hline
        Before Forget & 93.8 & 5.5 \\
        After Forget & 94.3 & 100 \\
        \hline
    \end{tabular}
    \label{clip_evaluation}
\end{table}

\end{document}